\documentclass{article}

\usepackage{arxiv}

\usepackage[utf8]{inputenc} % allow utf-8 input
\usepackage[T1]{fontenc}    % use 8-bit T1 fonts
\usepackage{hyperref}       % hyperlinks
\usepackage{url}            % simple URL typesetting
\usepackage{booktabs}       % professional-quality tables
\usepackage{amsfonts}       % blackboard math symbols
\usepackage{nicefrac}       % compact symbols for 1/2, etc.
\usepackage{microtype}      % microtypography
\usepackage{lipsum}
\usepackage{graphicx}
\usepackage{subfig}
\usepackage{setspace} 
\title{Hybrid Adaptive Neuro Fuzzy Inference System for Diagnosing the Liver Disorders }

\author{
  Mina Rajabi \\
  Department of Computer Science\\
  Yazd University\\
  \texttt{rajabi.mina@yazd.edu.ir} \\
   \And
 Hajar Sadeghizadeh  \\
  Department of Computer Science\\
  Islamic Azad University of Yazd\\
    \texttt{Sadeghizadeh@iau.yazd.edu.ir} \\
    \And
 Zahra Mola-Amini \\
  Department of Information Science\\
  Islamic Azad University of Meybod\\
    \texttt{MolaAmini@iau.meybod.edu.ir} \\
     \And
 Niloofar Ahmadyrad \\
  Department of Electrical Engineering\\
  Islamic Azad University of Meybod\\
    \texttt{Ahmadyrad@iau.meybod.edu.ir} \\
}
\begin{document}
\maketitle

\begin{abstract}
In this study, a hybrid method based on an Adaptive Neuro-Fuzzy Inference System (ANFIS) and Particle Swarm Optimization (PSO) for diagnosing Liver disorders (ANFIS-PSO) is introduced. This smart diagnosis method deals with a combination of making an inference system and optimization process which tries to tune the hyper-parameters of ANFIS based on the data-set. The Liver diseases characteristics are taken from the UCI Repository of Machine Learning Databases. The number of these characteristic attributes are 7, and the sample number is 354. The right diagnosis performance of the ANFIS-PSO intelligent medical system for liver disease is evaluated by using classification accuracy, sensitivity and specificity analysis, respectively. According to the experimental results, the performance of ANFIS-PSO can be more considerable than traditional FIS and ANFIS without optimization phase.

\end{abstract}

% keywords can be removed

\section{Introduction}
Expert systems are a branch of Artificial Intelligence (AI) which encourages unrestricted use of techno-scientific human expertise to solve semi or ill-structured issues wherever there is not a particular promise for determining the algorithm. The expert systems have been portrayed as an intelligent application that utilises knowledge and inference levels to manage severe problems to necessitate significant human expertise for their clarifications \cite{feigenbaum1981expert}.  

A fuzzy expert system can be a specific knowledge-based system, which is formed of fuzzification, knowledge database, inference rules,  and defuzzification parts, and applies fuzzy logic instead of the Boolean logic to consider about data in the deduction mechanism. This system is adopted to describe decision-making problems, where there is no scientific algorithm exists, although alternatively, the problem solution can be considered heuristically, which is based on specialists in the form of If-Then rules. A fuzzy expert system can be sufficiently supplied to the problem, which gives uncertainty emitting from fuzziness, ambiguity or subjectivity. In the 21st century, the application s of fuzzy expert systems have been tremendously expanding throughout scientific research topics such as diagnosing and predicting the various risk of the diseases \cite{polat2006diagnosis,csahan2007new,adeli2010fuzzy,neshat2015new,moya2019fuzzy}, civil engineering applications as an assistant \cite{neshat2011designing, neshat2012predication,yuan2014prediction,chiew2017fuzzy}, evaluating the educational service qualities \cite{pourahmad2012service,pourahmad2016using, raeesi2018quality,du2018fuzzy,shafii2016assessment} and for modelling different aspect of indeterministic business situations \cite{neshat2011fhesmm,neshat2016designing,rudzewicz2015quality,deveci2018interval}.   

The liver is one of the most significant body organs that detoxification of medications, elimination of reckless things emerging from the demolition and reconstruction of RBCs in the form of bile, composition of blood clotting parts, storage of sugar as glycogen. Moreover, the ordinance of sugar and fat metabolism are some of the essential functions of this body organ. We should not neglect its function in fat digestion and defence before microbus and toxins coming from foodstuff \cite{farokhzad2016novel}. In the recent decade, the death damage following various liver disorders has been dramatically expanding. On-time examination of this disease can be effective in the inhibition of its effects, its control and treatment. Browse considers expert's mentality as one of the most important issues in diagnosing disease because human-being is subject to error, and there is a possible error in disease diagnosis. One of the notable informatics medical procedures is to use expert systems to diagnose the disease with regard to a group of symptoms. These schemes can be based on artificial intelligence (AI) and assist experts to diagnose the diseases and more adequately satisfy them by acknowledging laboratory examinations. They also decrease cost,  save the time of experts and their incorrect judgment. Therefore, we have tried to diagnose liver disorders by using a hybrid adaptive neuro-fuzzy inference technique in this research. Our intentions are 1. choosing the ANFIS due to its advantages such as simple sense, high flexibility, the ability to endure fallacious data properly, modelling complex non- linear functions, to act on the basis of expert knowledge, the capability to conform with conventional controlling systems. 2. applying one of the most popular swarm intelligence technique called particle swarm optimization (PSO). PSO play a substantial impact to improve the performance of diagnosis by tuning the hyper-parameters of ANFIS like the number, type of fuzzy membership functions and enhancing the fuzzy rules.    

The main goal of this article is to address the important benefits and shortcomings of the current approaches and theories for improving and modelling fuzzy expert systems compared with the proposed hybrid adaptive neural fuzzy inference system performance for diagnosing the liver disorders based on the Bupa dataset. Concerning accomplishing like plans, a comprehensive study of the relevant fuzzy inference system technique is utilized.

The rest of this article is arranged as views. The details of the dataset used can be seen in Section 2. Section 3 illustrates the fuzzy expert systems in details.  In Section 4, the structure of adaptive neural fuzzy inference systems is reviewed. Besides, the technical specifications of the particle swarm optimization method and its diversity are studied in Section 5, and also section 6 shows the experimental implications. Conclusively, conclusions are sketched in Section 6.

%---------------------------------
\begin{figure}
\centering
  \includegraphics[width=0.6\linewidth]{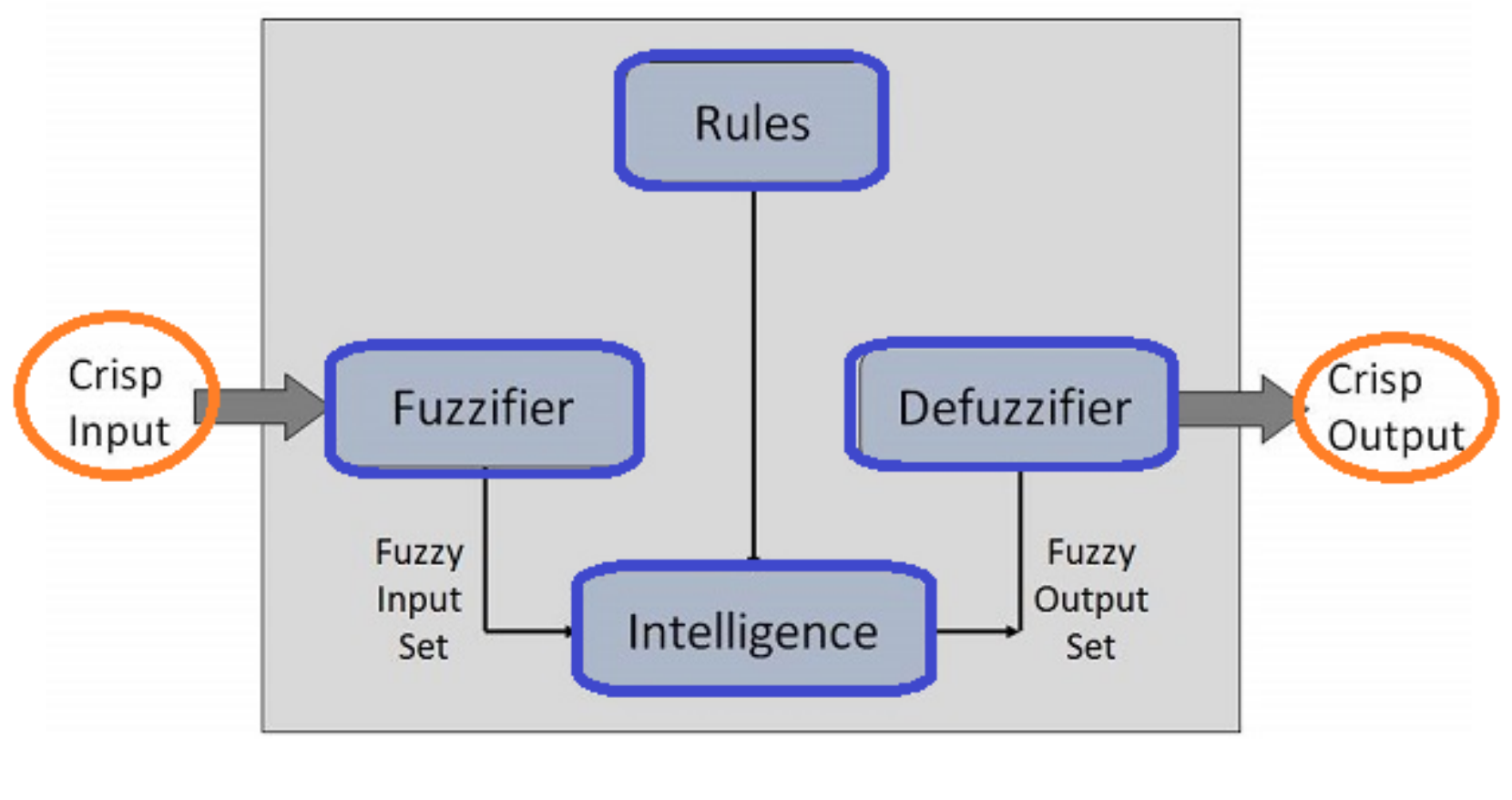}
  \caption{The scheme of a fuzzy expert system }
  \label{fig:FES}
\end{figure}
%--------------------------------------------------
\section{Bupa Liver Disorders Data-Set}
The dataset employed in this paper, which is used for enhancing the ability of liver disorders investigation according to their qualities, gathered by Richard s.forsyth and introduced to the UCI in 1995. The number of samples in this collection are 345, and each sample consists of 7 attributes.  In this dataset the initial five fields are related to variable substances of a male blood test, the 6th field is the quantity of
alcohol drinking, and lastly, the 7th field is using for restricting the healthy or ill individual. Attribute information can be seen in the following:\\
1.Mcv: means corpuscular volume Alkphos\\
2.Alkaline phosphates\\
3.Sgpt: alamine aminotransferase\\
4.Sgot: aspartate aminotransferase\\
5.Gammagt: gamma-glut amyl Tranpeptidase\\
6.Drinks: number of half-pint equivalents of
alcoholic beverages Drunk per day\\
7.Selector: field used to split data into two sets. \\
The dataset is continuous, and there is no missing or
destroyed data.

\section{Fuzzy Expert Systems (FES)}
\label{sec:fes}
Initially, Zadeh \cite{zadeh1996soft} introduced the main theory of fuzzy logic as an approach for interpreting human knowledge that is not precise and well-defined. Figure \ref{fig:FES} shows the fundamental form of a fuzzy logic system.  The process of fuzzification interface converts the crisp information into fuzzy linguistic values by various kinds of membership functions. The fuzzification can be regularly required in a fuzzy expert system considering the input values from surviving detectors are always deterministic numerical values. The inference generator demands fuzzy input and rules, and then it will produce fuzzy productions. Considerably, the fuzzy rule base should be in the figure of “IF-THEN” rules, including linguistic variables. The last part of a fuzzy expert system can be defuzzification which has the responsibility of performing crisp yield operations. The landscape of the fuzzy expert system can be represented in Figure \ref{fig:FES}.

In the last three decades, Fuzzy rule-based systems is a subsidiary of Artificial intelligence fitted of interpreting complicated medical data. Their potential to employ significant relationship within a data set has been used in the diagnosis, treatment and predicting consequence in various clinical outlines. A survey of different artificial intelligence methods is exhibited in this part, along with the study of critical clinical applications of expert systems. The ability of artificial intelligence systems and has been explored in almost every field of medicine. Artificial neural network and knowledge based systems were the most regularly accepted analytical tool while additional AI systems such as evolutionary algorithms, swarm intelligence and hybrid systems have been handled in various clinical environments. It can be concluded that AI and expert systems have a high potential to be employed in almost all fields of medicine. Table \ref{table:medicine} shows the application of practical AI techniques such as fuzzy sets, neural networks, evolutionary algorithms, swarm intelligence for diagnosing a wide set of diseases. Table \ref{table:medicine} shows a short review of different kinds methods for diagnosing the Liver disorders in the last two decades.  
%---------------------------------
\begin{figure}[h]
\centering
  \includegraphics[width=0.8\linewidth]{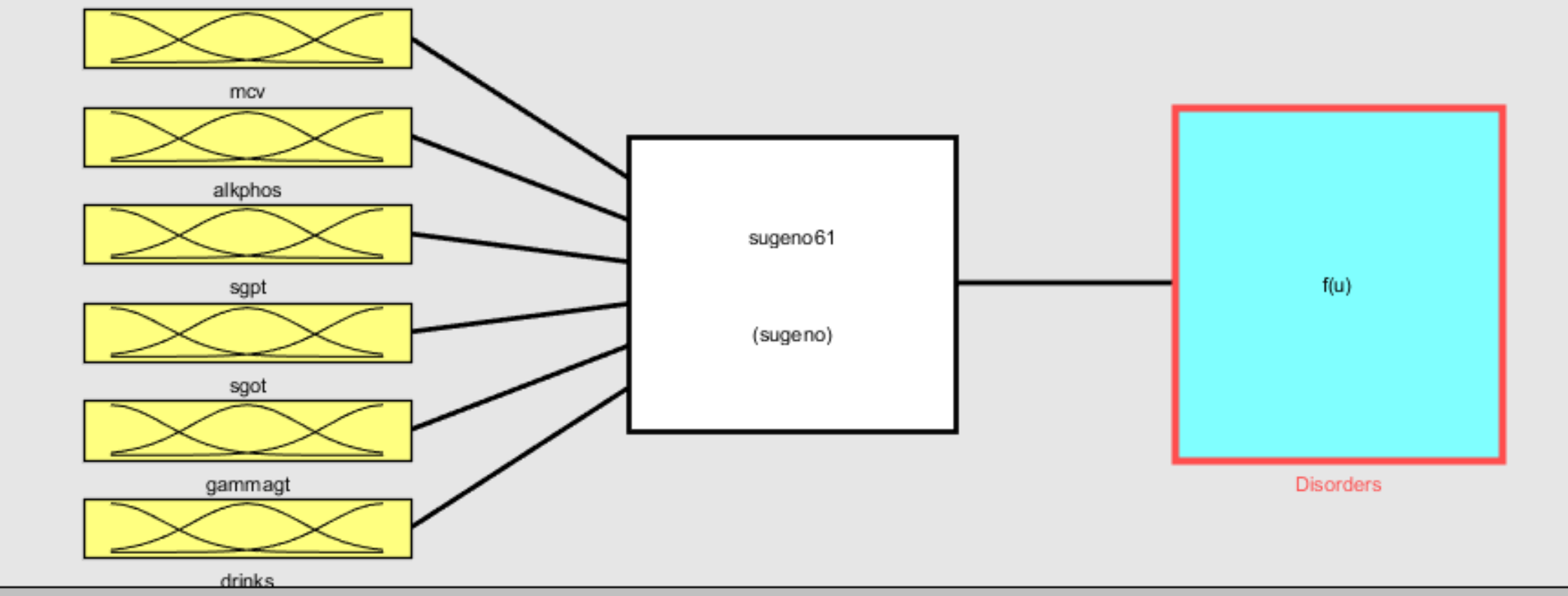}
  \caption{The scheme of the Liver disorders diagnotic Fuzzy Inference System}
  \label{fig:FIS}
\end{figure}
%--------------------------------------------------
%--------------------------------------
\begin{table}%[htbp]
\begin{center}
    \begin{tabular}{ p{3cm} | p{7cm} | p{3cm} | p{1cm} }
    \hline \hline
    \textbf{Authors} & \textbf{Methods} & \textbf{Disease} & \textbf{Year} \\ \hline
      Neshat et al. \cite{neshat2008designing,neshat2008fuzzy,neshat2010hopfield,neshat2014diagnosing,Neshat2013asurvey}& Bayesian parametric method and Parzen window non parametric method, Fuzzy Expert System, Hopfield Neural Network and Fuzzy Hopfield Neural Network & Liver Disease  &2008, 2009, 2010, 2013, 2014  \\ \hline
      Selvaraj et al.\cite{selvaraj2013improved}&particle swarm optimization  &Liver Disease &2013 \\ \hline
  Satarkar et al.\cite{satarkar2015fuzzy}&Fuzzy expert system &Liver Disease &2015 \\ \hline
    Hashemi et al. \cite{hashmi2015diagnosis} & fuzzy logic& Liver Disease &2015 \\ \hline
    Singh et al.\cite{singh2018efficient}&Principal Component Analysis and K-Nearest Neighbor (PCA-KNN) & Liver Disease &2018 \\ \hline
  Mirmozaffari et al. \cite{mirmozaffari2019developing}&expert system  & Liver Disease &2019 \\ \hline
   Kim et al. \cite{kim2014effective} & neural network and fuzzy neural network&Liver Cancer  & 2014 \\ \hline
  Das et al. \cite{das2018adaptive}&Adaptive fuzzy clustering-based texture analysis &Liver Cancer &2018 \\ \hline
   Xian et al. \cite{xian2010identification}& GLCM texture features and fuzzy SVM &Liver Tumors &2010 \\ \hline
    Polat et al. \cite{polat2007expert} & adaptive neuro-fuzzy inference system  & Diabetes Disease &2007  \\ \hline
    Polat et al. \cite{polat2006hepatitis}&artificial immune recognition system with fuzzy resource allocation  & Hepatitis Disease  & 2006 \\ \hline
    Chen et al.\cite{chen2011new}&local fisher discriminant analysis and support vector machines  & Hepatitis Disease &2011 \\ \hline
   Neshat et al. \cite{neshat2009designing, neshat2012hepatitis,neshat2009feshdd}& Adaptive Neural Network Fuzzy System, Hybrid Case Based Reasoning and PSO, Fuzzy expert system & Hepatitis B & 2009, 2012 \\ \hline
   Adeli et al. \cite{adeli2013new}&Genetic algorithm and adaptive network fuzzy inference system &Hepatitis  &2013 \\ \hline
  Ahmad et al. \cite{ahmad2018intelligent,ahmad2019automated}&adaptive neuro-fuzzy inference system, Multilayer Mamdani Fuzzy Inference System &Hepatitis Disease & 2018, 2019 \\ \hline
    \end{tabular}
    \caption{A briefly survey of the AI method applications for diagnosing the Liver disorders.}
\label{table:medicine}
\end{center}
\end{table}
%-----------------------------
----
\section{Adaptive Neural Fuzzy Inference System (ANFIS)}

\begin{figure}[h]
\centering
  \includegraphics[width=0.6\linewidth]{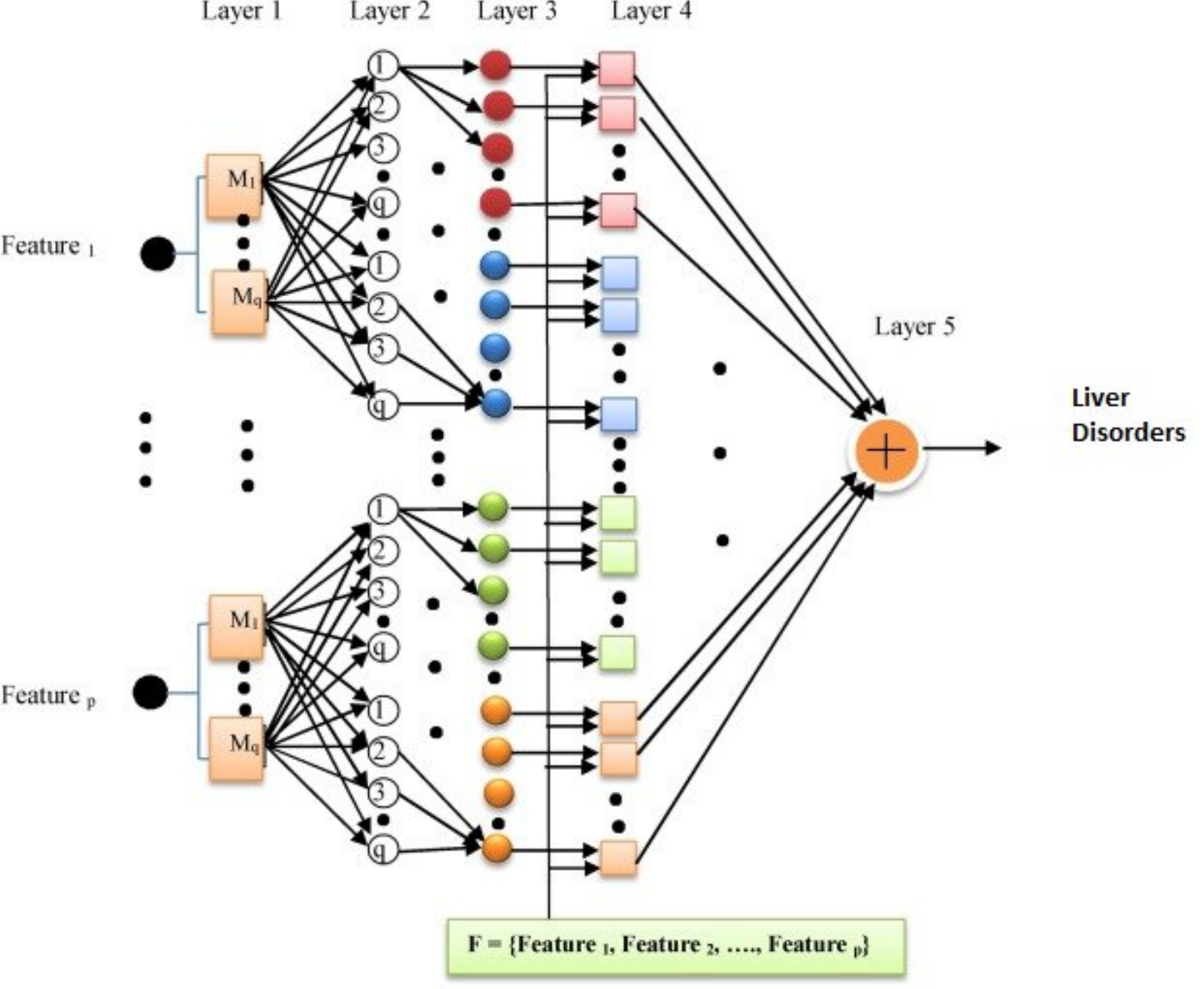}
  \caption{The five layers architecture  of an Adaptive Neuro-Fuzzy Inference System from \cite{nilashi2019predictive}}
  \label{fig:ANFIS}
\end{figure}

An adaptive neuro-fuzzy inference system (ANFIS) can be a class of artificial neural network (ANN) that is worked in regard to Takagi–Sugeno fuzzy inference system. The system was developed at the beginning of the 1990s \cite{jang1991fuzzy}. Since it combines both ANN and fuzzy logic principles, it holds the potential to catch the advantages of both in a unique framework. Its fuzzy inference system (FIS) corresponds to a collection of fuzzy rules (IF–THEN) which have learning inclination to approximate nonlinear functions. Consequently, ANFIS is supposed to be a general estimator. For practising the ANFIS more efficiently and optimally, one can handle the most useful parameters taken by genetic algorithm\cite{tahmasebi2012hybrid}. 
It is conceivable to distinguish two parts in the network structure, namely basis and consequence parts. In more details, the architecture is comprised of five layers. The first layer receives the input values and determines the membership functions referring to them. It is generally called fuzzification layer. The membership degrees of each function are calculated by applying the premise parameter set, namely {a,b,c}. The second layer is responsible for making the firing strengths for the rules. Due to its responsibility, the second layer is expressed as "rule layer". The role of the third layer is to normalize the measured firing strengths by diving each value for the total firing strength. The fourth layer practices as input the normalized values and the result parameter set {p,q,r}. The values yielded by this layer are the defuzzificated ones, and also those values are transferred to the last layer to replace the final output \cite{karaboga2018adaptive}. Figure \ref{fig:ANFIS} shows a deep landscape of ANFIS architecture. 
%---------------------------------
%----------------------------------------------

\section{Particle Swarm Optimization and its novelties}
Particle swarm optimisation (PSO) was introduced in 1995 by Kennedy and Eberhart \cite{trelea2003particle}, motivated by the operation of social animals in crowds, such as bird and fish schooling or ant colonies. This meta-heuristic follows the intercommunication among members to distribute knowledge. PSO has been implemented in diverse areas in optimisation and compound with other existing algorithms. This technique achieves the exploration of the optimal solution through particles, whose trajectories are modified by a stochastic and a deterministic element. Each particle is affected by its ‘best’ reached situation and the group ‘best’ situation, but leads to walking randomly. A particle $i$ is characterised by its status vector,$ x_i$, and its velocity vector,$ v_i$. Every repetition, each particle adjusts its situation according to the new velocity as:

%---------------------------------
\begin{figure}[h]
\centering
  \includegraphics[width=0.6\linewidth]{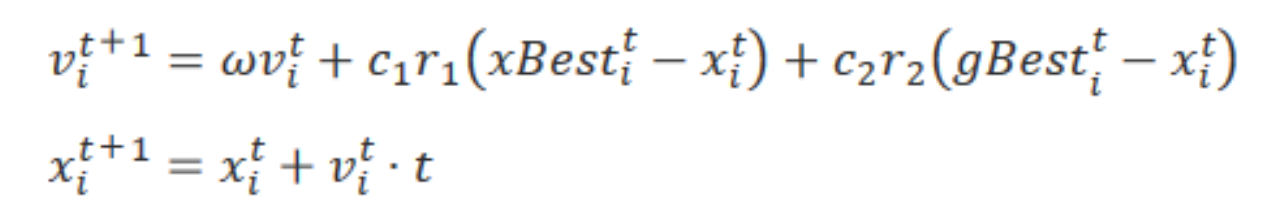}
  \end{figure}
%----------------------------------------------
where $xBest$ and $gBest$ indicate the best particle location and best group situation and the parameters $w, c_1, c_2, r_1$ and $r_2$ are respectively inertia weight, two positive constants and two random parameters within [0, 1]. In the baseline PSO, $w$ is chosen as a unit, but an enhancement of the PSO is observed in its inertial implementation using $w \in [0.5 0.9]$. Regularly, maximum and minimum velocity values are also described, and originally, the particles are assigned randomly to boost the search in all tolerable locations. Despite all the advantages of PSO like fast convergence and powerful in global search, its control parameters need to be tuned during the global search. There are plenty of proposed ideas to adjust the control parameters which have been called adaptive PSO \cite{zhan2009adaptive,neshat2010aipso,neshat2012new,hu2012adaptive,neshat2013faipso,lin2019adaptive}. Moreover, various meta-heuristic methods are combined with PSO to create a successful hybrid search technique \cite{naka2003hybrid,zahara2009hybrid,wang2018hybrid}. One of the benefits of PSO over other derivative-free approaches is the diminished the number of parameters to adjust and restrictions acceptance.    

\section{Hybrid Adaptive Neural Fuzzy Inference System (ANFIS-PSO)}
Development of the fancied fuzzy rule base can be a crucial step in
producing the fuzzy system. In prevailing, the rules and membership
function are made by specialists in a particular field, because the
meaning of these is commonly influenced by individual decisions.
While fuzzy rules denote comparatively straightforward to acquire by them, the MFs signify challenging to achieve. Tuning of MFs can be a time-consuming process. From the preceding analysis, it can be recognized that its membership function characterizes the fuzzy system and the type and parameters of determining the performance of the system MFs. Notwithstanding their attention, there are no observational techniques accessible for managing them. The fuzzy systems are formed as a search space, where each object in the space corresponds to a rule set and MFs. This performs evolutionary algorithms such as Genetic Algorithms (GAs), particle swarm optimization (PSO), better choices for
searching these spaces \cite{muthukaruppan2012hybrid}. Though PSO is similar to GAs, the principal difference between
them is that PSO is not equipped by genetic operators such as crossover
and mutation. Particles in PSO possess a memory which can be critical to the algorithm. When corresponded to GAs, the benefits of PSO are
purity in implementation and fewer parameters to adjust.

The steps of implementation of optimized FIS are in the following:

1. determining the test and train dataset

2. designing an initial FIS

3. Fine-tuning parameters of a model which must be adjusted very precisely in order to fit with the error function of the model by PSO.

4. Choosing the best FIS with the minimum RMSE as the best solution.

\section{Experimental outcomes}
\begin{figure}[t]
\subfloat[]{
\includegraphics[clip,width=0.49\columnwidth]{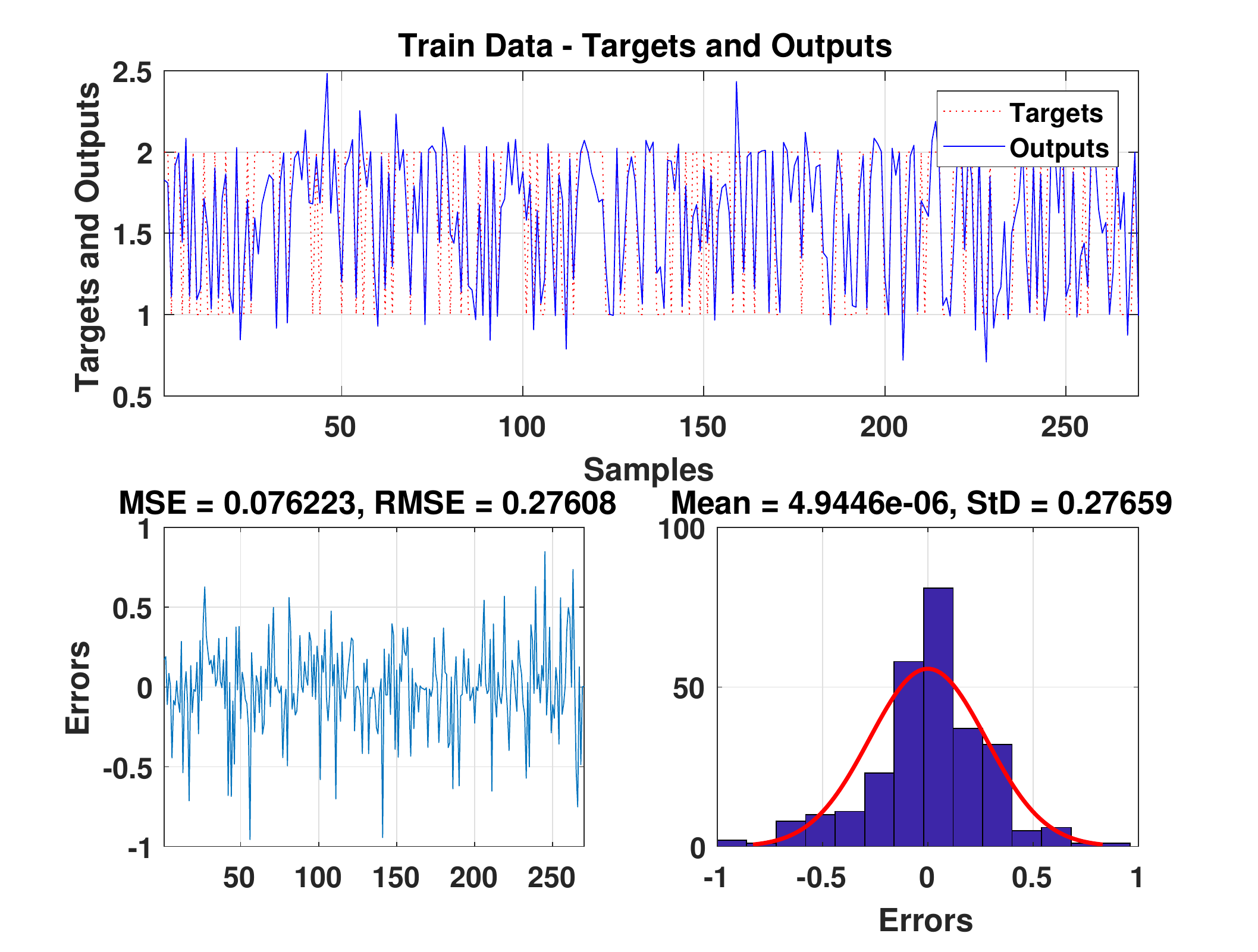}}
\subfloat[]{
\includegraphics[clip,width=0.49\columnwidth]{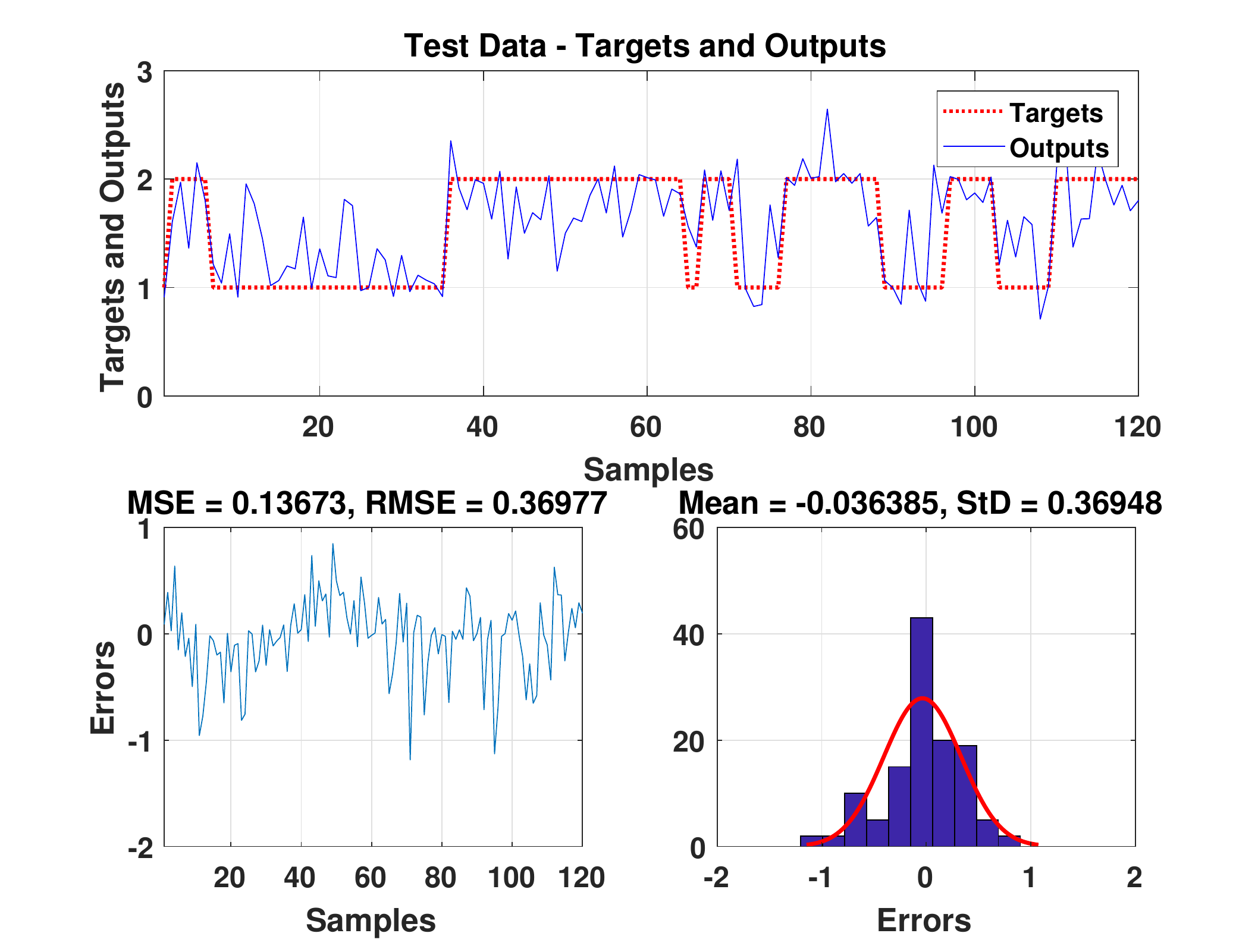}}
\caption{The training and testing results and error of an Adaptive Neuro-Fuzzy Inference System performance for diagnosing Liver disorders.}%
\label{fig:ANFIS_TEST_TRAIN}%
\end{figure}

\begin{figure}[h]
\centering
  \includegraphics[width=0.99\linewidth]{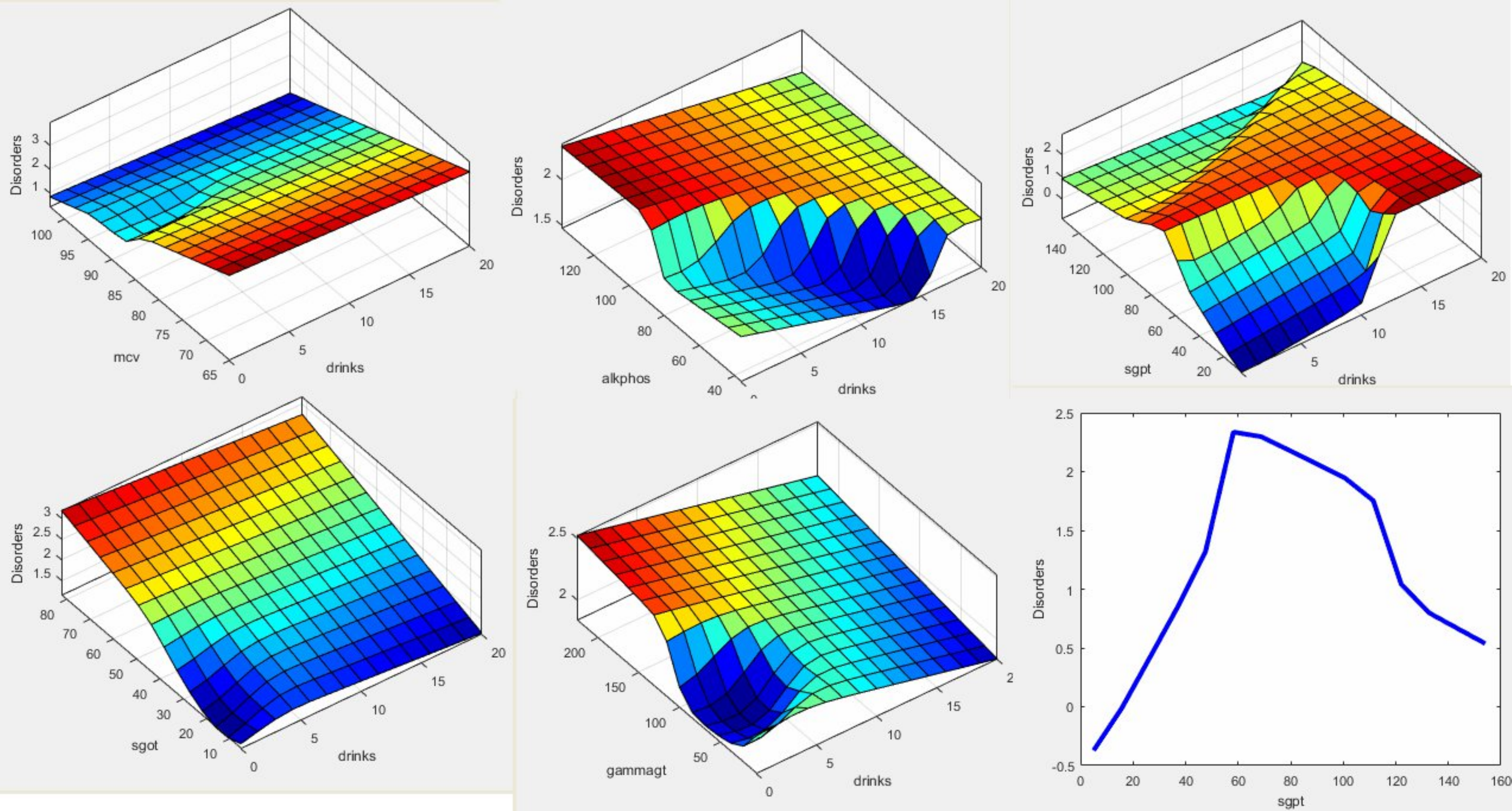}
  \caption{The fuzzy relationship among the effective variables of Liver disorders with different level of drink values BY ANFIS}%
  \label{fig:ANFIS_land}
  \end{figure}

\begin{figure}[h]
\centering
  \includegraphics[width=0.6\linewidth]{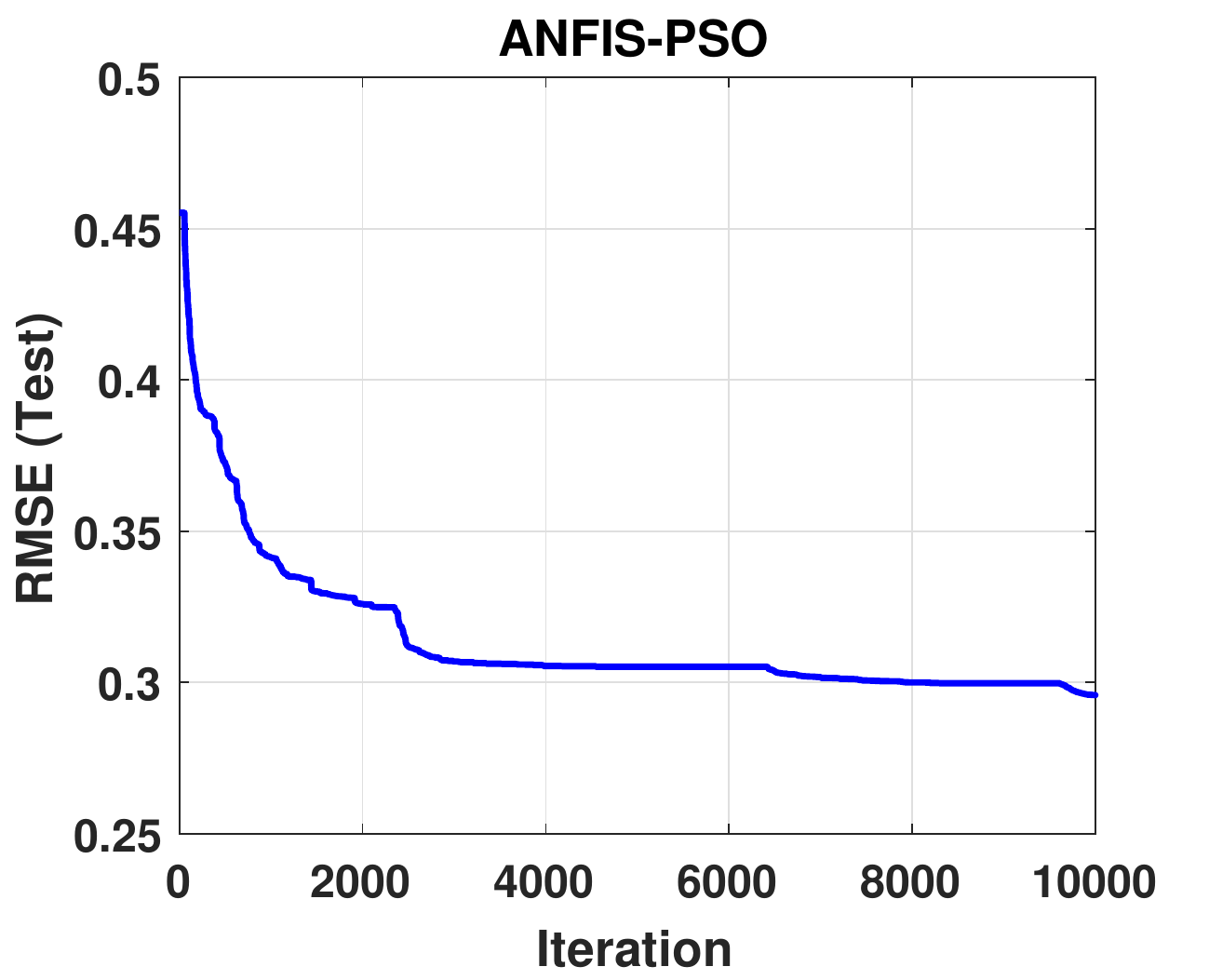}
  \caption{The performance of applied PSO to tune the FIS parameters and reduce the RMSE.}%
  \label{fig:ANFIS_PSO_per}
  \end{figure}

\begin{figure}[h]
\centering
  \includegraphics[width=0.99\linewidth]{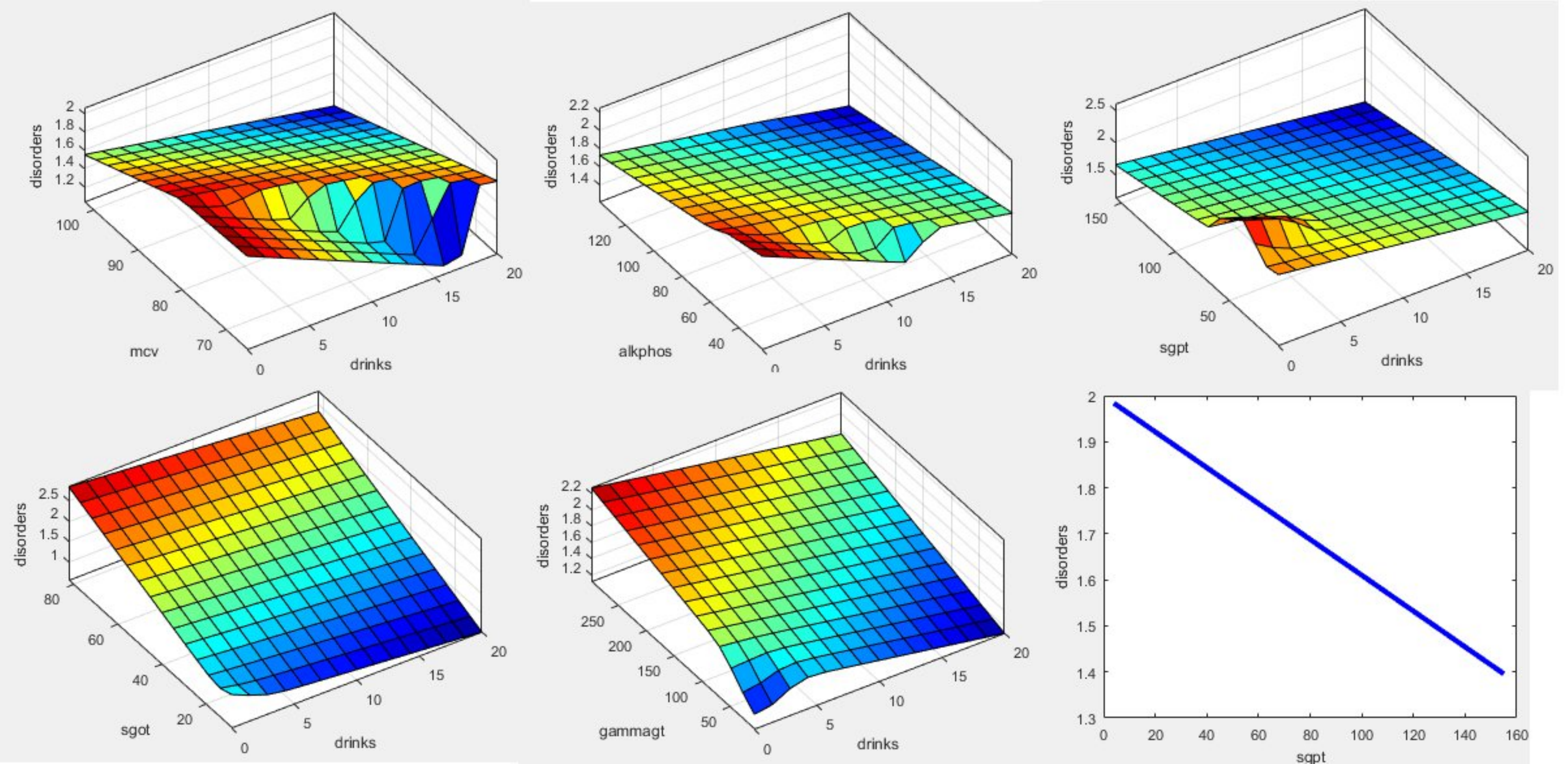}
  \caption{The fuzzy relationship among the effective variables of Liver disorders with different level of drink values BY ANFIS-PSO.}%
  \label{fig:ANFIS_PSO_land}
  \end{figure}

\begin{figure}[t]
\subfloat[]{
\includegraphics[clip,width=0.49\columnwidth]{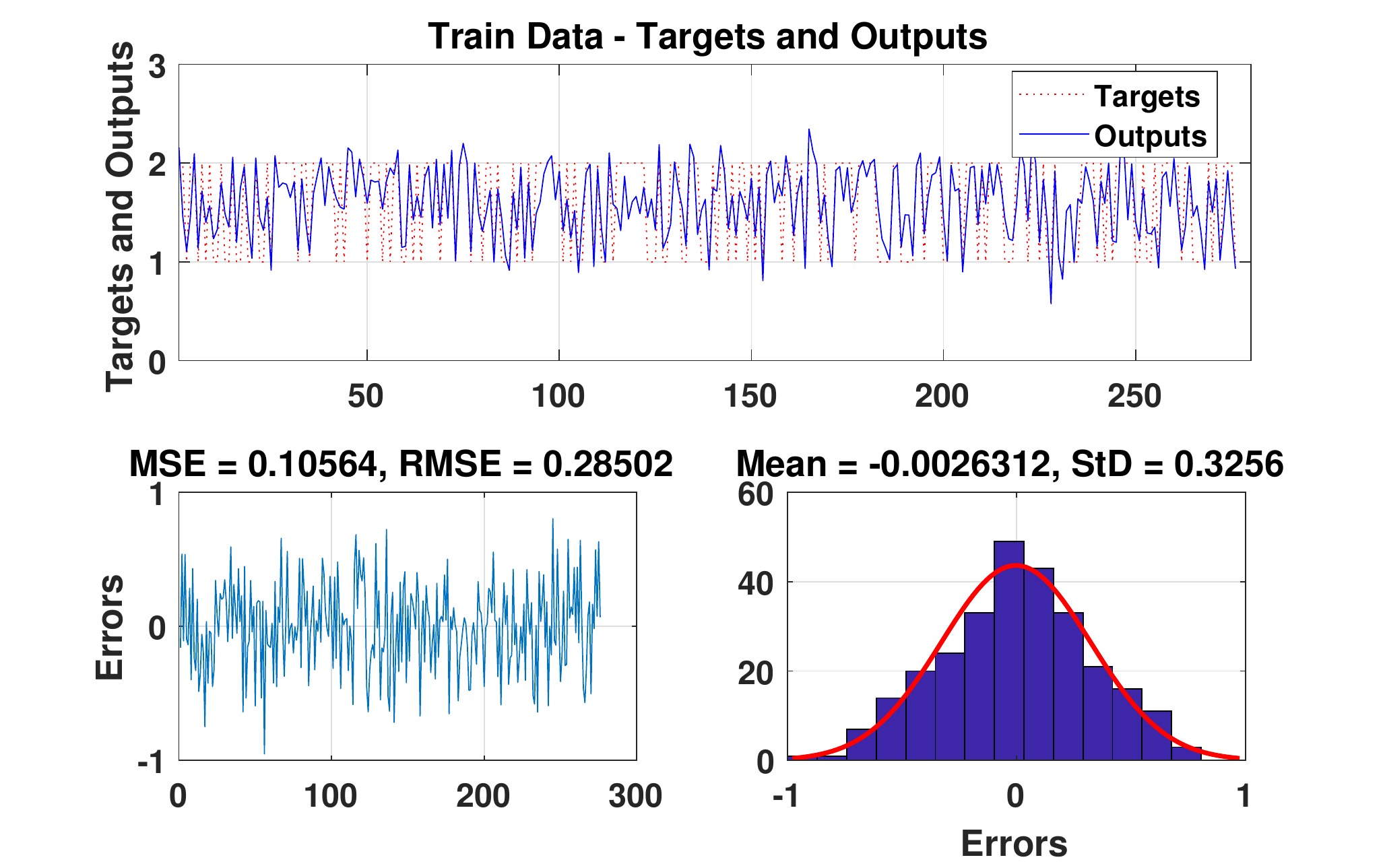}}
\subfloat[]{
\includegraphics[clip,width=0.49\columnwidth]{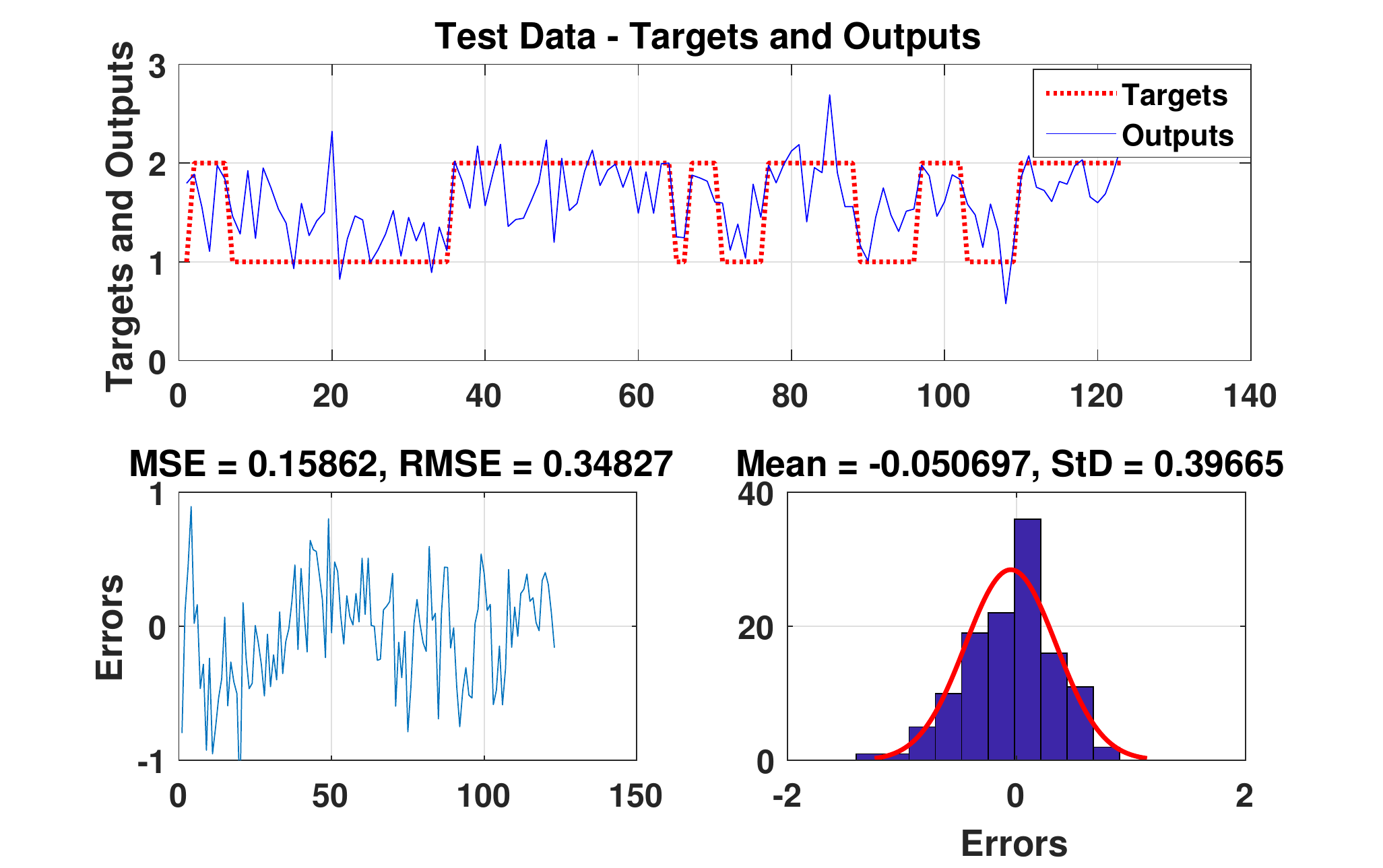}}
\caption{The training and testing results and error of an Adaptive Neuro-Fuzzy Inference System and PSO performance for diagnosing Liver disorders.}%
\label{fig:ANFIS_PSO_TEST_TRAIN}%
\end{figure}

The proposed ANFIS and ANFIS-PSO performances are evaluated by the dataset of Liver disorders (Bupa). The assumed evaluation criteria can be the mean square error (MSE) of the targets and predicted outputs. Figure \ref{fig:ANFIS_TEST_TRAIN} shows the MSE of both trained and tested data of ANFIS performance for one run. The distribution of error is regard to a normal distribution with an approximately wide variance which reveals that the used ANFIS requires some modifications.The average RMSE of ANFIS train and test are assigned at $0.292$ and $0.384$. On the other hand, we can see the improvement of the diagnosis accuracy by ANFIS-PSO in Figure \ref{fig:ANFIS_PSO_TEST_TRAIN}. The average RMSE of ANFIS-PSO training and testing are at $0.271$ and $0.343$. These results show an acceptable development for the new proposed method. Both Figure \ref{fig:ANFIS_land} and \ref{fig:ANFIS_PSO_land} present the fuzzy relationships among the 5 features with drinks level in ANFIS and ANFIS-PSO. It can be seen that the used ANFIS-PSO is able to enhance the achieved model. In the meantime, Figure \ref{fig:ANFIS_PSO_per} shows how PSO is able to tune the different parameters of used FIS.

% %--------------------------------
 % %----------------------------------------
%  % %--------------------------------
% \begin{figure}[t]
% \subfloat[]{
% \includegraphics[clip,width=0.33\columnwidth]{m1.png}}
% \subfloat[]{
% \includegraphics[clip,width=0.33\columnwidth]{m2.png}}
% \subfloat[]{
% \includegraphics[clip,width=0.33\columnwidth]{m3.png}}\\
% \subfloat[]{
% \includegraphics[clip,width=0.33\columnwidth]{m4.png}}
% \subfloat[]{
% \includegraphics[clip,width=0.33\columnwidth]{m5.png}}
% \subfloat[]{
% \includegraphics[clip,width=0.33\columnwidth]{m6.png}}
% \caption{The fuzzy relationship among the effective variables of Liver disorders with different level of drink values BY ANFIS.}%
% \label{fig:ANFIS_land}%
% \end{figure}
%  % %---------------------------------------
 %---------------------------------
%----------------------------------------------
 %---------------------------------
%----------------------------------------------
%---------------------------------
%-----------------------------------------
% %--------------------------------
 % %----------------------------------------

\section{  Conclusions}
\label{sec:con}
In this study, a hybrid adaptive neural fuzzy expert system based on particle swarm optimization (PSO) was revealed in Matlab’s Simulink in order to distinguish Liver disease and health condition. With this recommended strategy, the accuracy of  classification is increased by $10\%$ compared with the ANFIS based on the dataset can be accomplished. The development of the meaningful attributes and fuzzy rules were obtained using the statistical analysis. The importance of identifying significant and relevant fuzzy rules without the assistance of the specialists exposes the potentiality of knowledge discovery. The principal benefits of the FIS as a knowledge acquisition mechanism are the following: (1) adaptive number of rules are concerned (2) the acquired rules can be efficiently explained. These results propose encouraging research areas employing PSO and fuzzy expert system in several classification problems. According to the achieved results, the hybrid proposed system is able to beat previous studied approaches in terms of both accuracy and reliability.

\bibliographystyle{unsrt}  
\bibliography{references}

\end{document}